\documentclass[12pt]{article}
\usepackage[margin=1in]{geometry}
\usepackage{setspace}
\usepackage{amsmath,amssymb,amsthm}
\usepackage{graphicx,psfrag,epsf}
\usepackage{natbib}
\usepackage{url} 
\usepackage{booktabs}
\usepackage{hyperref}
\usepackage{float}
\usepackage{enumerate}
\usepackage{adjustbox}
\usepackage{diagbox}
\usepackage{subfigure}
\usepackage{cleveref}

\theoremstyle{plain}

\theoremstyle{definition}

\theoremstyle{remark}

\title{Probabilistic Functional Neural Networks}
\author{Haixu Wang\thanks{Department of Mathematics and Statistics, University of Calgary, Calgary, Canada.} 
\and Jiguo Cao\thanks{Department of Statistics and Actuarial Science, Simon Fraser University, Burnaby, Canada. }}

\date{}

\begin{document}

\maketitle

\begin{abstract}
High-dimensional functional time series (HDFTS) are often characterized by nonlinear trends and high spatial dimensions. Such data poses unique challenges for modeling and forecasting due to the nonlinearity, nonstationarity, and high dimensionality. We propose a novel probabilistic functional neural network (\textit{ProFnet}) to address these challenges. \textit{ProFnet} integrates the strengths of feedforward and deep neural networks with probabilistic modeling. The model generates probabilistic forecasts using Monte Carlo sampling and also enables the quantification of uncertainty in predictions. While capturing both temporal and spatial dependencies across multiple regions, \textit{ProFnet} offers a scalable and unified solution for large datasets. Applications to Japan’s mortality rates demonstrate superior performance. This approach enhances predictive accuracy and provides interpretable uncertainty estimates, making it a valuable tool for forecasting complex high-dimensional functional data and HDFTS.
\end{abstract}


\section{Introduction}
High-dimensional functional time series (HDFTS) have emerged as a unique data structure in fields such as econometrics, public health, and environmental science. This data type involves observations of functional data, like curves or surfaces, across multiple regions over time. Analyzing and forecasting HDFTS presents unique challenges due to its inherent nonlinearity, nonstationarity, and high dimensionality, where the number of spatial units often exceeds temporal observations.

\begin{figure}[H]
    \centering
    \includegraphics[width=\textwidth]{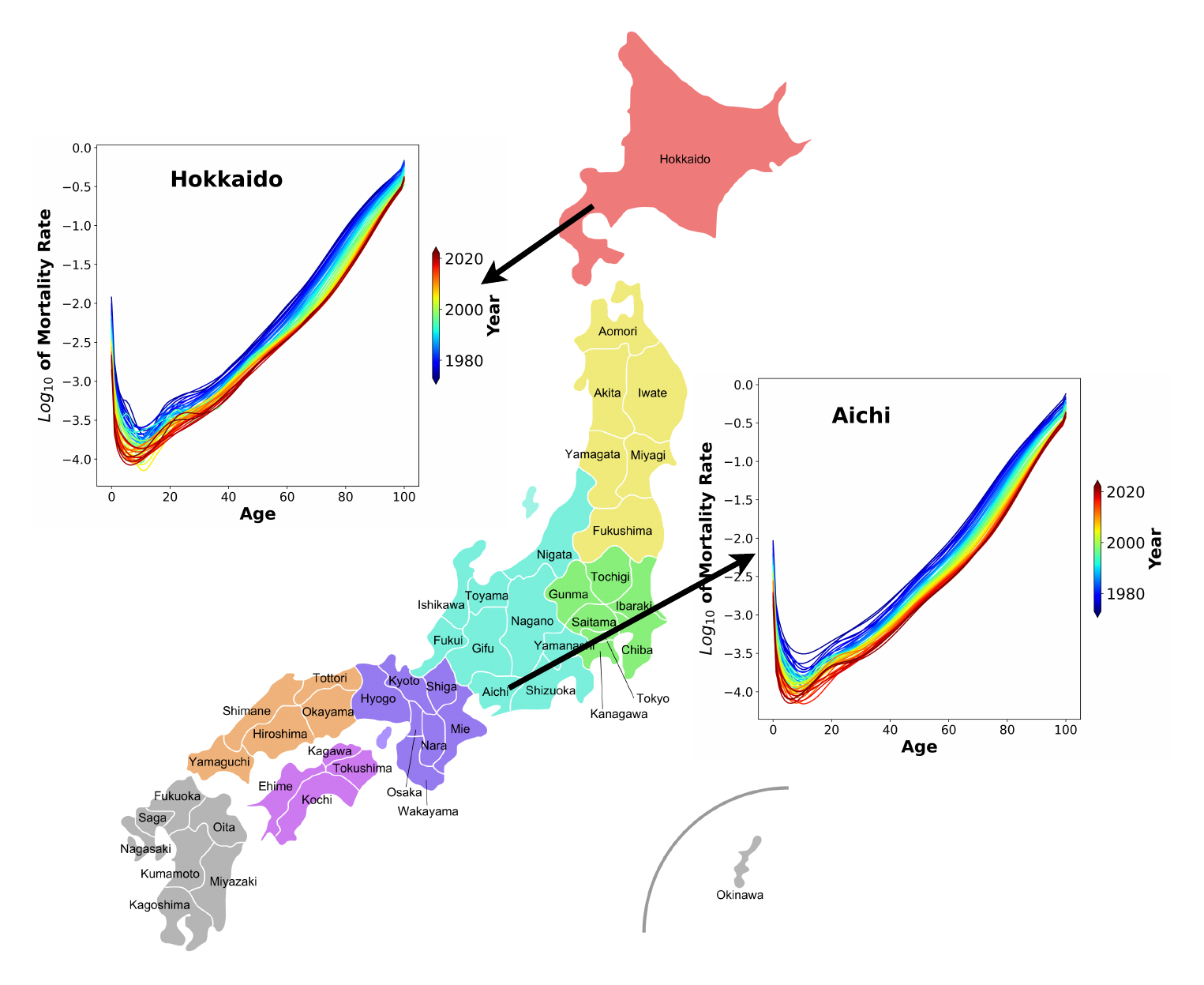}
    \caption{\label{fig:data_intro} An example of high-dimensional functional time series. The data is a set of mortality rate curves for 47 prefectures in Japan from year 1973 to 2022. The color represents the year of the mortality rate curve, from blue (oldest) to red (most current). We would simultaneously observe a functional time series within each prefecture, and the collection of all prefectures forms a high-dimensional functional time series. The plot showcases the functional time series for 2 randomly chosen prefectures.}
\end{figure}

For instance, consider monitoring mortality rate curves across Japan's 47 prefectures from 1973 to 2022 as in Figure~\ref{fig:data_intro}. Each prefecture contributes a time series of functional data, and collectively, these form a high-dimensional dataset. This emerging type of functional data urges a unified and flexible method that should be robust and scalable.

Analyzing high-dimensional functional time series (HDFTS) often extends from established methods for univariate functional time series \cite{funseries_score1,funseries_score2}. Traditional analyses typically fall into two categories: functional principal component analysis (FPCA) and functional regression models. The first approach focus on predicting future functional data by obtaining their projections on the estimated functional principal components, e.g., \citet{JAO_partialFTS, inferenceforfunctionaldatawithapplications, sparsefts, meanoffts}. 

This approach assumes the stationarity and linearity of the data and heavily depends on lag-specific forecasting. Functional regression models, on the other hand, adopt a regression-based framework in which past functional observations serve as covariates to predict future data. This methodology, similar to a functional counterpart of autoregressive models, has been applied in studies like \cite{funseries_pls} and \cite{functionalautoregression}. A nonparametric model has been discussed in \cite{projection_FTS}. Although these models address some of the shortcomings of FPCA, they also rely on the same rigid assumptions of stationarity and linearity, limiting their applicability in real-world scenarios characterized by complex and non-stationary trends. 

Conventional FPCA focuses on univariate functional data and struggles with multivariate or high-dimensional settings. A common workaround is to concatenate multiple dimensions of functional data into a single functional object and apply univariate FPCA techniques \cite{fdabook}. While effective to some extent, this approach is computationally expensive and overlooks the discontinuous connecting points between functional variables. Some alternative FPCA methods are developed for multivariate functional data \citet{chiouFPCA,mfpca_differentdomain,PCAformultivariate}. 

In recent years, attention has turned to explicitly tackling HDFTS. Some important work has been developed by \citet{GSY19, LLS24, JSS24}. In addition, \citet{ZD23} explored statistical inference, \citet{TSY22} investigated clustering, and \citet{HNT231} employed factor models for forecasting. To tack non-stationarity, \citet{dynamicFPCA} and \citet{dynamicFPCA_freshwater} employed a dynamic version of FPCA for forecasting. Despite these advancements, many remain direct extensions of univariate methods and fail to fully address the limitations of stationarity, linearity, and lag dependence.

An alternative approach to modeling and forecasting high-dimensional functional time series (HDFTS) is the application of neural networks. Among these, recurrent neural networks (RNNs) have gained popularity for their ability to model temporal dependencies in time series data. RNN-based strategies have been extensively reviewed and discussed, highlighting their strengths and limitations in various forecasting contexts \cite{HEWAMALAGE2021388}. Recent developments, such as the robust forecasting model proposed in \citet{ZHANG2023143} and the sequence-to-sequence framework introduced in \citet{seq2seq}, have enhanced the utility of RNNs in handling complex temporal structures. For functional data, RNNs have been adapted to forecast functional time series, as demonstrated in \citet{NOP}. These models are highly effective at capturing nonlinearity and temporal patterns but face several significant challenges. First, RNN-based methods are typically lag-dependent, requiring a fixed network architecture specifically designed for certain time lags. Additionally, they often encounter difficulties during training and incur high computational costs, making them less suitable for HDFTS. Finally, a critical limitation of RNNs in the context of HDFTS is their inability to quantify forecast uncertainty. In many applications, particularly in areas such as public health and finance, uncertainty quantification is as important as the point forecast itself for enabling informed decision-making and robust statistical inference.


In summary, HDFTS poses a complex data structure that demands flexible and scalable models for effective analysis and forecasting. Existing methodologies suffer from limitations of linearity and stationarity assumptions, and lack of scalability or uncertainty quantification. In this work, we propose a probabilistic functional neural network (\textit{ProFnet}) for analyzing and forecasting HDFTS. To the best of our knowledge, this is the first attempt to build a probabilistic model and address the uncertainty of functional data with neural networks. We outline the key contributions to the forecasting problem and advantages over traditional functional data analysis (FDA) methods and RNN-based models:

\noindent\textbf{Probabilistic forecasting.} The foremost contribution of this work is to build a probabilistic model for HDFTS. In fact, it can be seen as a generative model quantifying the forecast uncertainty through Monte Carlo samples and consequently generated functional data. Then, we can quantify the uncertainty through the distribution of generated functional data. This capability is essential for applications where reliable predictions and uncertainty estimates can support more informed decision-making. Compared to the existing approaches, \textit{ProFnet} simultaneously addresses the complexity and nonlinearity of functional data while combining the strengths of neural networks and statistical inference.

\noindent\textbf{Joint modeling of temporal and spatial dependencies.} In this work, we will focus on the application with dimensionality corresponding to the spatial dimension. The proposed method simultaneously captures temporal and spatial dependencies across regions, modeling how functional data for one region is influenced by its own history and that of other regions. This simultaneous modeling of all regions offers a significant advantage over traditional FDA methods, which often require fixed designs and individual models for each region. \textit{ProFnet}'s flexibility allows it to be adapted to other spatial and multivariate scenarios.

\noindent\textbf{Scalability and flexibility.} Built on a feedforward neural network structure, \textit{ProFnet} is inherently scalable and better suited to high-dimensional and large-scale datasets than RNN-based models. It accommodates trends and nonstationarity in functional data, overcoming the limitations of traditional FDA approaches. Furthermore, \textit{ProFnet} supports forecasting across multiple time lags simultaneously, enhancing efficiency and accuracy, whereas both existing statistical and RNN models are typically lag-dependent. 

The rest of the paper is organized as follows. In Section~\ref{sec:method}, we introduce the proposed probabilistic neural network \textit{ProFnet} for high-dimensional functional time series. Section~\ref{sec:application} presents the application of the proposed method in forecasting Japan's mortality rate curves, and we compare the method with some existing frameworks based on either machine learning or traditional statistical models. In Section~\ref{sec:simulation}, we conduct simulation studies to evaluate the forecast performances of the proposed model as well as sensitivity analysis on its scalability. Finally, we conclude the paper in Section~\ref{sec:conclusion}.

\section{Method}\label{sec:method}


Let $X_{t,h}(u)$ represent the individual functional data at the time point $t$ and in a sub-region $h$ for $t = 1,..., T$ and $h = 1,,,. H$. Every individual curve $X_{t,h}(u) \in \mathcal{L}_{2}(\tau)$ is a square-integrable function over a compact support $u \in \tau$. In many HDFTS settings, the number of regions ($H$) often exceeds the number of time points ($T$), making this a high-dimensional data problem. The primary objective of \textit{ProFnet} is to construct a unified model that forecasts all regions $h$ simultaneously, avoiding the inefficiency and redundancy of building separate models for each region.


\subsection{Probabilistic Functional Neural Network}

\textit{ProFnet} integrates probabilistic modeling with deep learning to capture temporal and spatial dependencies while quantifying forecast uncertainty. It has three key components: Encoding, Probabilistic Block, and Generator. 

\noindent\textbf{Functional and Spatial Encoding.} The model takes the functional data $X_{t,h}(u)$ and the region $h$ as inputs. We start with representation learning of the input. That is, we introduce two mappings $\phi_{x}(\cdot) : \mathcal{L}_{2}(\tau) \rightarrow \mathbb{R}^{L_{x}}$ and $\phi_{h}(\cdot): \mathbb{N} \rightarrow \mathbb{R}^{L_{h}}$ to encode the functional data and region, respectively. The former is a mapping on functions, and the latter is a mapping for integer-valued indicators for all regions $h$. Hence, the functional data is encoded into a latent representation $\mathbf{W}_{x} = \phi_{x}(X_{t,h}(u))$ and the region is encoded into a latent representation $\mathbf{W}_{h} = \phi_{h}(h)$. 

The representation learning on functional data, learning $\phi_{x}(\cdot)$, can be done in several ways, such as using a functional linear model, functional principal component analysis, or discrete versions of dimension reduction technique. In this work, we consider a functional version of a feedforward neural network to obtain the latent representation of individual functional data. Consider a 1-layer (functional learning layer) network taking the following form:
\begin{align}
    \mathbf{W}_{x} = \sigma([&\langle \beta_{1}(t), X_{t,h}(u)\rangle_{\mathcal{H}}, ...,  \nonumber\\
                            &\langle \beta_{L_{x}}(t), X_{t,h}(u)\rangle_{\mathcal{H}}]) \label{eq:functional_layer}
\end{align}
with the feedforward operation $\sigma$ (inner-product and non-linear activation) and giving an output of $L_{x}$ dimensional latent vector $\mathbf{W}_{x}$. The norm $\langle \cdot, \cdot \rangle_{\mathcal{H}}$ is the inner product in the Hilbert space $\mathcal{H}$ of the functional data and takes the form $\langle f, g \rangle_{\mathcal{H}} = \int_{u \in \tau} f(u)g(u) du$. The parameters of this layer is the collection of functional coefficients $\beta_{l}(t)$'s for $l = 1,...,L_{x}$. 

To facilitate the training of the neural network, we can interpret $\beta_{l}(t)$ as a linear combination of known or pre-fixed basis functions, i.e.,  $\beta_{l}(t) = \sum_{d=1}^{D}\omega_{dl}\phi_{d}(t)$. We can see that the inner product in Equation~\ref{eq:functional_layer} can be computed as $\langle \beta_{l}(t), X_{t,h}(u)\rangle_{\mathcal{H}} = \sum_{d=1}^{D}\omega_{dl}\int_{u}\phi_{d}(t)X_{t,h}(u) du$, and it becomes a multiple linear regression problem on $\omega_{dl}$'s which can be integrated to the backward and forward operations of standard neural network. Some common choices of basis functions are B-spline, Fourier basis, or the empirical basis functions obtained from the functional principal component analysis. 

To add depth of $\phi_{x}(\cdot)$ in learning the latent representation of $X_{t,h}(u)$, we can stack traditional fully connected layers to the functional learning layer in Equation~\ref{eq:functional_layer}. That is, we can define a functional learning block as:
\begin{equation}
    \mathbf{W}_{x} = \sigma( \underbrace{\cdots}_{\text{J-1}} \sigma([\langle \beta_{1}(t), X_{t,h}(u)\rangle_{\mathcal{H}}, ..., ]))
    \label{eq:functional_block}
\end{equation} 
with $J-1$ fully connected layers, one functional learning layer, and the feedforward operation $\sigma$. The input functional data can be either discretized or continuous. The former requires us to approximate any integrals with the Riemann sum. The latter usually have the functional input data already in a basis expansion form which can be directly incorporated into the functional learning layer. 

Here, we refer to $\phi_{h}(\cdot)$ as the spatial learning block. The latent representation of $h$ can obtained by using an embedding layer or a matrix factorization scheme. Assuming we are using a 1-layer learning block for $h$, the former is a lookup table that maps the region $h$ to a latent vector $\mathbf{W}_{h}$. The latter is a matrix factorization scheme that factorizes the region matrix $Q$ (calculated based on pair-wise distances) into two low-rank matrices $\mathbf{G}$ and $\mathbf{V}$ such that $\mathbf{Q} \approx \mathbf{G} \mathbf{V}^{T}$. The latent representation of $h$ is the $h-th$ row vector of $G$, $\mathbf{G}_{h\cdot}$, corresponding to the region $h$. We can add more layers to the initial learning layer to obtain the latent representation of $h$ in a deep learning fashion. Let $W_{h,1}$ be the initially learned representation of $h$ through either using network embedding layer or matrix factorization, we can obtain the final latent representation of $h$ as:
\begin{equation}
    \mathbf{W}_{h} = \sigma( \underbrace{\cdots}_{\text{J-1}} \sigma(W_{h,1}))
    \label{eq:spatial_block}
\end{equation}
which defines our spatial learning block with $J$ layers as in Equation~\ref{eq:spatial_block}. Here, we are using the same $J$ for simplicity of notations. $J$ could be differnt for either Equation~\ref{eq:functional_block} or~\ref{eq:spatial_block}. All operations in the spatial learning block are compatible with a feedforward neural network and can be trained with a backpropagation algorithm.

After combining the two learning blocks, we obtain the final (nonlinear and potentially deep) latent representation $\mathbf{W} = [\mathbf{W}_{x}, \mathbf{W}_{h}]$ for a given input pair $(X_{t,h}(u), h)$. The probabilistic component will be built based on $\mathbf{W} \in \mathbb{R}^{L = L_{x} + L_{h}}$ rather than the unprocessed functional inputs. 

\noindent\textbf{Probabilistic block.} The probabilistic block is a collection of Gaussian processes (GPs) as functions of $t$, and they are parametrized and conditioned on the functional and spatial inputs through their latent representation $\mathbf{W}$. The Gaussian processes are then used to model the distribution of the functional data $X_{t,h}(u)$ for all $t$ and $h$. We let $z_{k}(t)$ be the random sample from the $k$-th Gaussian process $GP_{k}(t)$ at time $t$ for $k = 1,..., K$. Each Gaussian process is parameterized with a mean function $m_{k}(t)$ and covariance function $c_{k}(t,t{^\prime})$. Each pair $m_{k}(t)$ and $c_{k}(t,t{^\prime})$ can be further parameterized or represented with more complex models, i.e., combination or product of different mean and kernel components. This is beyond the scope of this paper, and we can consult \cite{GPML,kernelcooking} for more advanced options. For simplicity, we will use a constant mean function and squared exponential kernel for illustration. Both the mean and covariance components of $K$ Gaussian processes will be functions of $\mathbf{W}$ and implemented with feedforward layers. That is, we could define our parameter learning block as follows:
\begin{align}
    m_{k}(t) &\equiv \mu:  \mu = \sigma( \underbrace{\cdots}_{\text{J}} \sigma(\mathbf{W})) \nonumber \\ 
    c_{k}(t,t^{\prime}) &\propto \exp(-\frac{(t - t^{\prime})^2}{2\rho^{2}}): \rho = \sigma( \underbrace{\cdots}_{\text{J}} \sigma(\mathbf{W}))
    \label{eq:probabilistic_learning}
\end{align}
where $\sigma$ represents the feedforward operation in neural networks as a combination of inner-product and nonlinear activation. The learning block, denoted as $\gamma(\mathbf{W}) = [\gamma_{\mu}(\mathbf{W}), \gamma_{c}(\mathbf{W})]$ in the future, in Equation~\ref{eq:probabilistic_learning} can be applied to more advanced choice of mean functions and kernels, and it has two components: mean learning $\gamma_{\mu}(\mathbf{W})$ and kernel learning $\gamma_{c}(\mathbf{W})$. Here, we assume all $K$ Gaussian processes are independent of each other. As a result, the probabilistic block in Equation~\ref{eq:prob_block} has an output of $z_{k}(t^{\prime})$:
\begin{equation}
   z_{k}(t^{\prime}) \sim N(\mu_{k}(t^{\prime}|t,\mathbf{W}), c_{k}(t^{\prime}|t,\mathbf{W}))
\label{eq:prob_block}
\end{equation}
for $k = 1,..., K$.

Assuming we want to forecast the functional time series at a future time point $t^{\prime}$. We consider the distribution of a pair of random points $(z_{k}(t), z_{k}(t^{\prime}))$ which are jointly Gaussian. Given a random sample $z_{k}(t)$, the conditional distribution of $z_{k}(t^{\prime})$ is Gaussian with mean $m_{k}(t^{\prime}) + c_{k}(t^{\prime},t) c_{k}(t,t)^{-1}(z_{k}(t) - m_{k}(t))$ and variance $c_{k}(t^{\prime},t^{\prime}) - c_{k}(t^{\prime},t) c_{k}(t,t)^{-1} c_{k}(t,t^{\prime})$. We can sample from the conditional distribution of $z_{k}(t^{\prime})$ given $z_{k}(t)$ as in Equation~\ref{eq:prob_block}, then we repeat this process for all $K$ Gaussian processes to generate $K$ random samples $\mathbf{z}(t^{\prime}) = (z_{1}(t^{\prime}),...,z_{K}(t^{\prime}))^{T}$.

In this work, we focus on the inference from $t$ to $t^{\prime}$ where the region information is not used in the probabilistic block but added later. This allows us to identify the deterministic effect or region-to-region relationship and provides more interpretable results. The network architecture is illustrated in the following Figure~\ref{fig:FunProbNetwork}.
\begin{figure}[H]
    \centering
    \includegraphics[width=\textwidth]{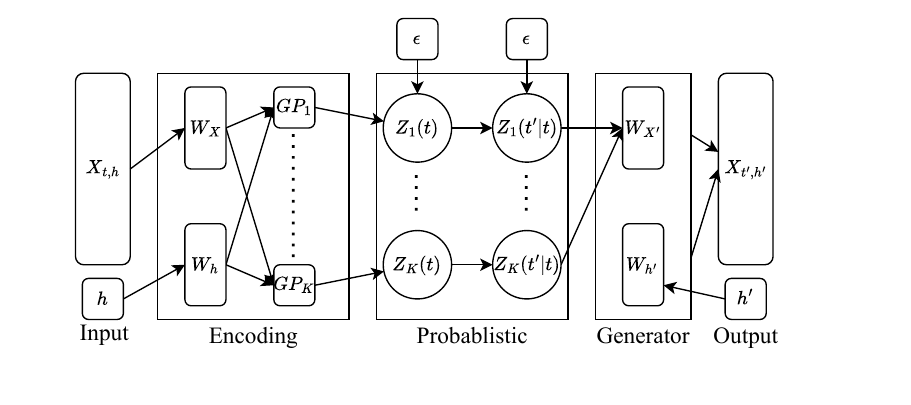}
    \caption{Functional Probabilistic Neural Network. Arrows indicate the forward pass of the network. The reparametrization trick is done by adding pre-sampled random noises in $\mathbf{\epsilon}$.}
    \label{fig:FunProbNetwork}
\end{figure}

\noindent\textbf{Generator.} The generator block is a generative model as a consequence of the probabilistic block. For each sampled $\mathbf{z}(t^{\prime})$ with deterministic $\mathbf{W}_{h^{\prime}}$, we can generate a new functional data $\hat{X}_{t^{\prime},h^{\prime}}(u)$. Similar to the functional learning block in Equation~\ref{eq:functional_block}, the generator block can directly generate a discretized curve or a continuous output. As a common practice, we let the generator first produce a latent representation vector then the actual forecast of $X_{t^{\prime},h^{\prime}}(u)$. 

We can define a functional generator block $\psi_{x}(\cdot)$ taking the input of $\mathbf{W}^{\prime} = [\mathbf{W}_{x^{\prime}},\mathbf{W}_{h^{\prime}}]$, where 
\begin{equation*}
    \mathbf{W}_{x^{\prime}} = \sigma( \cdots \sigma( \sigma( \mathbf{z}(t^{\prime}) ) ) )
\end{equation*}
where $\sigma$ is the feedforward operation in fully connected layers and $W_{h^{\prime}}$ is the region information obtained by using the previous spatial learning block in Equation~\ref{eq:spatial_block}. The output of the generator block is $\hat{X}_{t^{\prime},h^{\prime}}(u) = \psi_{x}(\mathbf{W}^{\prime})$ where $\psi_{x}$ serves as an inverse of the functional learning layer in Equation~\ref{eq:functional_layer}.


The pseudocode of a forward pass for the proposed \textit{ProFnet} is summarized as follows:
\begin{itemize}
\item Input: $X_{t,h}(u)$ and $h$; Target $t^{\prime}$ and $h^{\prime}$.
\item Encoding: 
\begin{itemize}
    \item Latent representations with funtional and spatial learning blocks $\phi_{x}(\cdot)$ and $\phi_{h}(\cdot)$: $W_{x} = \phi_{x}(X_{t,h}(u))$ and $W_{h} = \phi_{h}(h)$. The latent representations are concatenated to form $W = [W_{x}, W_{h}]$.
    \item Consider a collection of $K$ Gaussian process defined on the support of $t$ for $X_{t,h}(u)$ with the parameter set $\{\mu_{k},\rho_{k}\}_{k=1}^{K}$. We use parameter learning block in Equation~\ref{eq:probabilistic_learning} to obtain the parameters $\{\mu_{k},\rho_{k}\}_{k=1}^{K}$ based on the latent representation $W$.
\end{itemize}
\item Sampling: 
\begin{itemize}
    \item For each Gaussian process, we generate a pair of random samples $z_{k}(t), z_{k}(t^{\prime})$ from a bivariate normal distribution. Where the latter (posterior) is conditioned on the former (prior).
    \item Concatenate all $K$ random samples $\mathbf{z}(t^{\prime}) = (z_{1}(t^{\prime}),...,z_{K}(t^{\prime}))$.
\end{itemize}
\item Generator: 
    \begin{itemize}
        \item Use the conditional random samples $\mathbf{z}(t^{\prime})$ to explain the latent representation of target functional data $W_{x^{\prime}}$.
        \item Concatenate latent representation of target region $W_{h^{\prime}}$ and previously obtained $W_{x^{\prime}}$, we can generate the final output, the forecasted functional data $\hat{X}_{t^{\prime},h^{\prime}}(u) = \psi_{x}([W_{h^{\prime}}, W_{x^{\prime}}])$.
    \end{itemize}
\end{itemize}

\subsection{Training}
We will refer to the functional generator block as a distribution of functions in $\mathcal{L}_{2}(\tau)$, i.e., $P(X_{t,h}(u)|\mathbf{Z}(t);\mathbf{\omega},W_{h})$ where $\mathbf{\omega}$ contains all the fixed neural network parameters involved in the generator block. This is to align the notations used in  \cite{kingmavae2019}. Assume that all $X_{t,h}(u)$'s are evaluated at the same set of grid points $u_{1},..., u_{M}$, and this is only for simplicity and not a necessary condition. We let $\mathbf{X}_{t,h} = (X_{t,h}(u_{1}), ..., X_{t,h}(u_{M}))$ and choose $P(\mathbf{X}_{t,h}|\mathbf{Z}(t);\mathbf{\omega},W_{h})$ to be a multivariate Gaussian distribution with mean $\mathbf{\mu}_{t,h} = f(\mathbf{z}(t);\mathbf{\omega},W_{h})$ and a diagonal covariance matrix $I_{M}$. The deterministic function $f(\cdot)$, an umbrella function including operations on $\mathbf{z}(t)$ and $\psi_{x}(\cdot)$, is the output of the generator block and including all involved neural network layers. The first component of the objective function is the negative log-likelihood which can be equivalently represented by the mean square error (MSE) loss, i.e.,
\begin{equation}
    \mathcal{L} = -\text{log}P(X_{t,h}(u)|\mathbf{Z}(t);\mathbf{\omega},W_{h}) \propto \frac{1}{M}|| \mathbf{X}_{t,h} - \mathbf{\mu}_{t,h} ||^2
    \label{eq:reconloss}
\end{equation}
where $||\cdot||$ is the Euclidean norm. In the proposed model, we are trying to obtain a variational distribution $Q(\mathbf{z}(t^{\prime})|X_{t,h}(u))$ rather than the true posterior distribution $P(\mathbf{z}(t^{\prime})|X_{t,h}(u))$. The variational distribution is parameterized by a neural network $Q(\mathbf{Z}(t^{\prime})|X_{t,h}(u);\mathbf{\theta})$ with parameters $\mathbf{\theta}$ involved in all layers within the encoding block. The second component of the objective function is the Kullback-Leibler (KL) divergence between the variational distribution and the prior distribution $P(\mathbf{Z}(t^{\prime}))$. The KL divergence is defined as 
\begin{align}
    &\text{KL}(Q(\mathbf{Z}(t^{\prime})|X_{t,h}(u))||P(\mathbf{Z}(t^{\prime}))) \nonumber \\ 
    &= \mathbb{E}_{Q}\log [Q(\mathbf{Z}(t^{\prime})|X_{t,h}(u)) -  P(\mathbf{Z}(t^{\prime}))]
    \label{eq:KL-divergence}
\end{align}
with $P(\mathbf{Z}(t^{\prime}))$ being the prior distribution of $K$ Gaussian processes at $t^{\prime}$. After putting the two components together, the objective function becomes

\begin{align}
    \mathbb{E}_{Q} \{\frac{1}{M}|| \mathbf{X}_{t^{\prime},h^{\prime}} &- \mathbf{\mu}_{t^{\prime},h^{\prime}} ||^2  + \nonumber\\
    &\log [Q(\mathbf{Z}(t^{\prime})|X_{t,h}(u)) -  P(\mathbf{Z}(t^{\prime}))]\} \nonumber \nonumber\\
     \approx \frac{1}{M}|| \mathbf{X}_{t^{\prime},h^{\prime}} &- \mathbf{\mu}_{t^{\prime},h^{\prime}} ||^2    + \nonumber\\
     &\log [Q(\mathbf{z}(t^{\prime})|X_{t,h}(u)) -  P(\mathbf{z}(t^{\prime}))]
    \label{eq:objective}
\end{align}
with a single sampled $\mathbf{z}(t^{\prime})$ from $Q(\mathbf{Z}(t^{\prime})|X_{t,h}(u))$. The objective function in Equation~\ref{eq:objective} is the loss function for the proposed \textit{ProFnet} and can be optimized with the stochastic gradient descent algorithm given one pair of $(X_{t,h}(u), X_{t^{\prime},h^{\prime}}(u))$. 

The proposed \textit{ProFnet} is a flexible model that can be trained similarly to a feed-forward network, hence it ensures scalability and efficiency. It also has two major advantages over either traditional FDA-based approaches or RNN-structured neural networks. First, the proposed network is a lag-free framework. Existing forecast models are typically lag-dependent, meaning they are designed for specific pairs of time points $t$ and $t^{\prime}$, often expressed as $t$ and $t + \delta$ with $\delta$ being fixed. The $\textit{ProFnet}$ allows to be trained with any pair of $t$ and $t^{\prime} \geq t$, and it can be trained with multiple pairs of $t$ and $t^{\prime}$ simultaneously for any pair of regions $(h,h)$ or $(h, h^{\prime})$. This ensures better generalization and increases data usability. Second, \textit{ProFnet} is particularly well-suited for cases where the number of regions ($H$) is much larger than the number of time points ($T$). By jointly modeling all regions and lags, \textit{ProFnet} mitigates the challenges of limited temporal data and over-parameterization often encountered in deep neural networks.

\section{Applications}\label{sec:application}

The application focus on studying a nation's mortality rate which is a public health indicator. The rate is defined to be the probability of death at a particular age in a population. A nation's mortality is recorded along with region-specific mortality information, such as the prefectures in Japan. In this application, we wish to use the \textit{ProFnet} to forecast Japan's mortality rate curves for all prefectures and a set of forecast horizons simultaneously rather than building individual models. The data set is a collection of smoothed $\log_{10}$ mortality rate curves from year 1973 to 2022 for all 47 prefectures in Japan. The data set is split into training and testing data sets with a ratio of 80\% and 20\%, respectively. The training data set is used to train the \textit{ProFnet} and the testing data set is used to evaluate the prediction performance.

\begin{figure}[H]
    \centering
    \includegraphics[width=\textwidth]{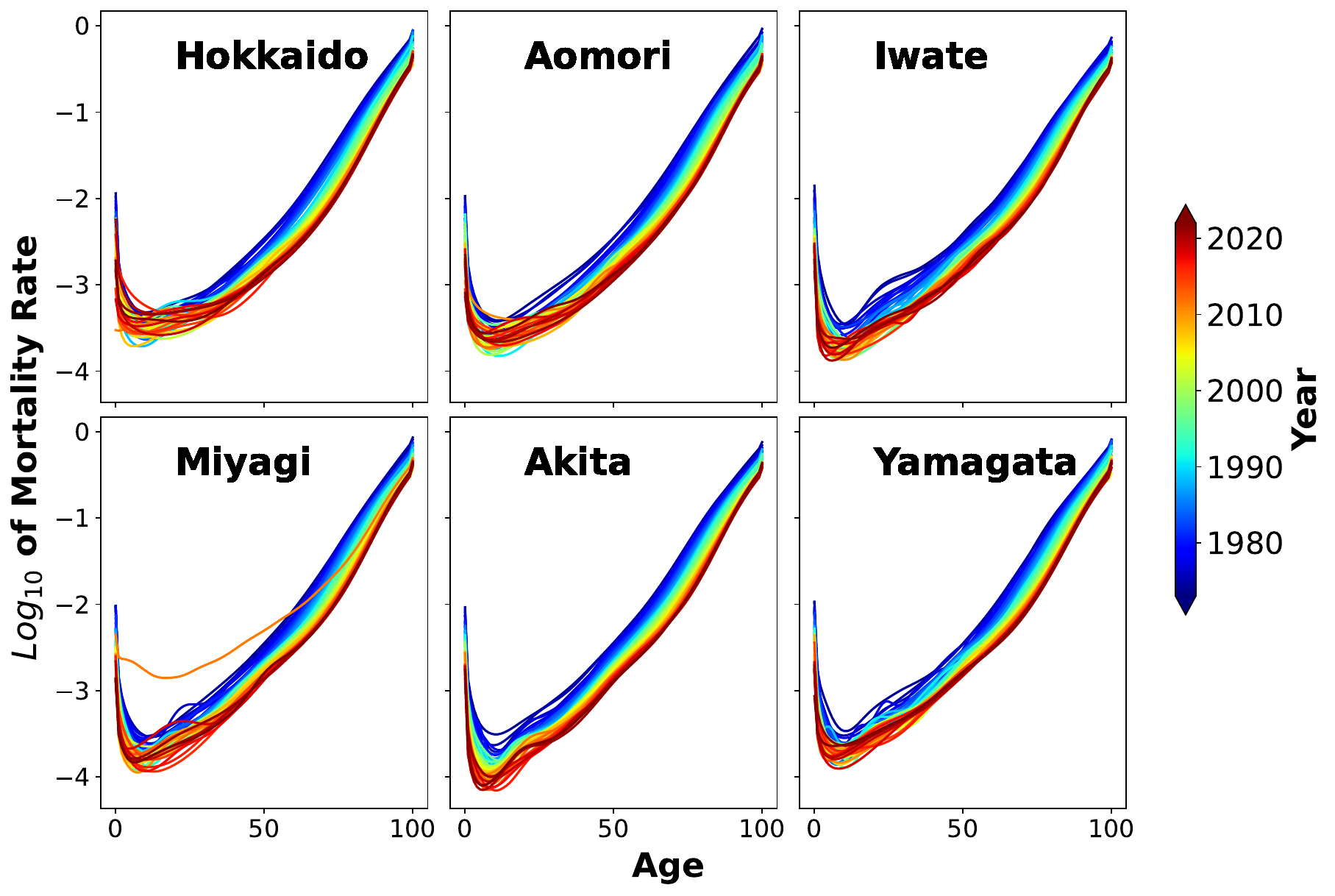}
    \caption{\label{fig:data_preview} Smoothed $\log_{10}$ mortality rate curves, from year 1973 to 2022, for six randomly chosen prefectures (with its name in the plot) in Japan. The color represents the year of the mortality rate curve, from blue (oldest) to red (most current).}
    \end{figure}

From Figure~\ref{fig:data_preview}, we can see a general decreasing trend in the mortality rate curve as time goes from past to more recent years. This trend is visible in all prefectures of Japan. In the meantime, we can also observe differences in the mortality rate across sub-regions. For example, the mortality rate of Iwate seems to be higher than the other prefectures, and the trend is not as clearly observed as in other prefectures. We could observe a similar functional time series for the Akita prefecture. For the prefectures of Aomori and Yamagata, we could observe a dip in the mortality rate for the age group of 8-16. One thing to notice is that we will not detrend the data as a common practice in time series analysis. The reason is that the trend is an important part of the data and we want to capture the trend in the forecast model. This also demonstrates the flexibility of our proposed model in handling the trend and nonstationarity of high-dimensional functional time series.

When we are comparing the forecasted curves in the testing data, we use the mean square forecast error (MSFE) which is defined to be:   
\begin{equation*}
    \text{MSFE}_{h} = \frac{1}{T^{\prime}} \sum_{t^{\prime}=1}^{T^{\prime}}\int\big[X_{t,h}(u) - \widehat{X}_{t,h}(u)\big]^2 du,
    \end{equation*}
for all $T^{\prime}$ unobserved curves in the testing data set and a given region $h$. We will compare the MSFE of the proposed \textit{ProFnet} with the functional bivariate linear model (FLM), the nonlinear prediction method (NOP in \cite{NOP}), the univariate functional time series (UFTS), the multivariate functional time series (MFTS), and the multivariate functional linear time series (MFLTS). The comparison is done for different forecast horizons $\delta = 1, 5, 10$, and first we will use $\text{MSFE} = \sum_{h=1}^{H}\text{MSFE}_{h}$ to evaluate the overall forecast performance. The results are summarized in Table~\ref{tab:1}. Here, the ProFnet is trained based on a fixed forecast horizon $\delta$ and the point estimate for future functional data is obtained by taking one random sample from all latent Gaussian processes. The $\text{ProFnet}_{\text{mean}}$ is to use the generator to generate multiple samples and take the point-wise (at a single $u$) average of generated functional data as a forecast.
\begin{table}[t]
    \caption{\label{tab:1} The prediction performances MSFE of different methods in forecasting the mortality rate curves in testing data for one step ahead and averaged over all prefectures.}
    \begin{center}
    \begin{small}
    \begin{sc}
    \begin{tabular}{lccccl}
    \toprule
    \backslashbox{Method}{Lag} & $\delta = 1$ & $\delta = 5$ & $\delta = 10$ \\
    \midrule
    $\text{ProFnet}_{\text{mean}}$ & 0.011 & 0.012 & 0.009 \\
    ProFnet & 0.019 & 0.023 & 0.025 \\
    FLM & 0.053 & 0.053 & 0.053 \\
    NOP & 0.056 & 0.056 & 0.056 \\
    UFTS & 0.053 & 0.053 & 0.053 \\
    MFTS& 0.058 & 0.058 & 0.058 \\
    MFLTS& 0.054 & 0.054 & 0.054 \\  
\bottomrule
\end{tabular}
\end{sc}
\end{small}
\end{center}
\vskip -0.1in
\end{table}
We can see that the proposed neural network is superior in forecasting future data. In addition, it is more computationally feasible and scalable to high-dimensional functional time series data compared to the existing NOP method. The latter is built on a recurrent neural network. \textit{ProbFnet} takes around 2 minutes to train with stochastic gradient descent whereas \textit{NOP} usually takes more than 30 minutes.

\begin{figure}[H]
    \centering
    \includegraphics[width=\textwidth]{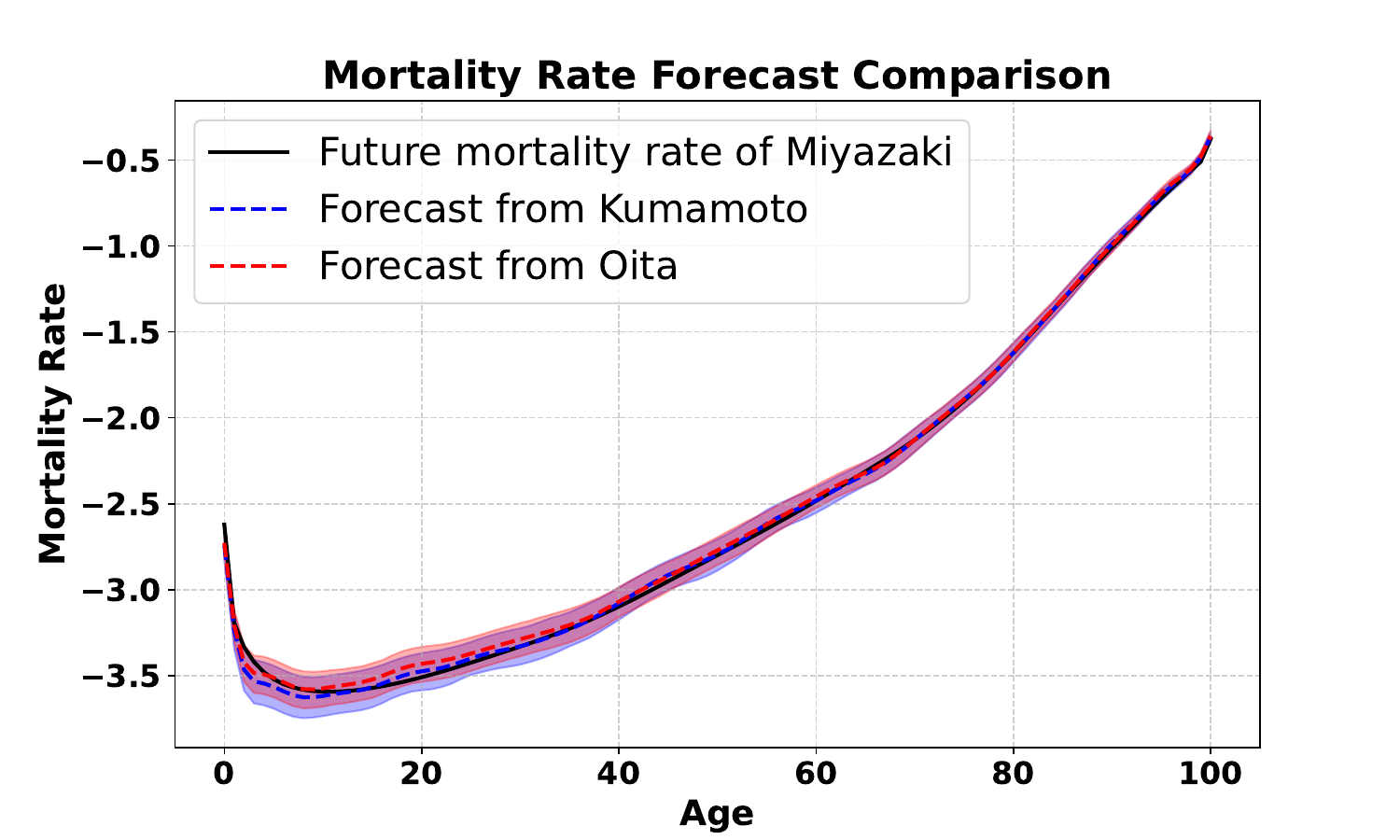}
    \caption{\label{fig:intervalestimate} Forecast results for a target region, Miyazaki prefecture. The interval estimates are obtained by using regional data from Kumamoto or Oita. The true future curve is the solid black line, whereas the blue dashed line is the mean estimate from Kumamoto and the red dashed line is that from Oita. The shaded area represents the 95\% forecast interval.}
\end{figure}

Since we have proposed a probabilistic model, a single point estimate (in the sense of functional data, a point estimate is a single curve) is less favorable than an interval estimate (an interval estimate in the functional data case is to see how the forecasted curve varies along the support of function). The interval estimate can be obtained by generating random values from the collection of Gaussian processes and using them to generate samples of the target future functional data. The interval estimate, for example in Figure~\ref{fig:intervalestimate}, is more informative and can quantify uncertainty, and we can also use the generated functional data to calculate a coverage probability which is defined as the proportion of the true functional data within the $(1-\alpha)\%$ prediction interval. That is, $(1-\alpha)\%$ prediction interval is calculated based on 
\begin{equation*}
    Q(\frac{\alpha}{2};u) = \text{quantile}(\frac{\alpha}{2};\{\hat{X}_{t^{\prime},h^{\prime}}(u)\})
\end{equation*}
where $\{\hat{X}_{t^{\prime},h^{\prime}}(u)\}$ is the collection of generated functional data at $u$.

We then calculate these quantiles at $u_{1},..., u_{M}$ where we have observed the true functional data. The coverage probability is then calculated as
\begin{equation*}
\frac{1}{M}\sum_{m=1}^{M} \mathbb{I}(X_{t^{\prime},h^{\prime}}(u_{m}) \in [Q(\frac{\alpha}{2};u_{m}), Q(1-\frac{\alpha}{2};u_{m})])
\end{equation*}
We have summarized the coverage probability of the forecasted curve in Table~\ref{tab:2} for different forecast horizons. 
\begin{table}[H]
    \caption{\label{tab:2} The mean coverage probability of the proposed \textit{ProFnet} over all prefectures, i.e., for each target prefecture we take the maximum coverage probability and the average of them over the total number of prefectures.}
     \begin{center}
    \begin{small}
    \begin{sc}
    \begin{tabular}{lccccl}
    \toprule
    Lag & $\delta = 1$ & $\delta = 5$ & $\delta = 10$ \\
    \midrule
    Coverage & 98.3\%   & 95.8\%  & 85.6\%  \\
\bottomrule
\end{tabular}
\end{sc}
\end{small}
\end{center}
\vskip -0.1in
\end{table}

The proposed \textit{ProFnet} demonstrates consistently high coverage probabilities across all forecast horizons. This result highlights \textit{ProFnet}'ss ability to deliver reliable forecast intervals for future functional data. Notably, the coverage probability tends to decrease with larger forecast horizons, aligning with expectations as forecast uncertainty naturally grows over time.

One benefit of the proposed \textit{ProFnet} is that we can have interpretable regional associations, and the associations have directions through the coverage probabilities. For example in Figure~\ref{fig:intervalestimate}, we are able to get two sets (should be $H$ sets for the entire data) of estimates for a single target region.

The forecast from Kumamoto is better than that from Oita based on the coverage. Hence, we can assume a stronger directional association from Okinawa to Kagoshima. We can then plot the entire directional associations obtained by doing forecasting against the associations by ranking the geographical distances between regions. The result is illustrated in the following Figure~\ref{fig:connection}. The distance-based associations do not present any interpretable information. However, the connections derived from the coverage probabilities are more interpretable and can be used to study the regional associations of the mortality rates in different regions.

\begin{figure}[H]
    \centering
    \includegraphics[width=\textwidth]{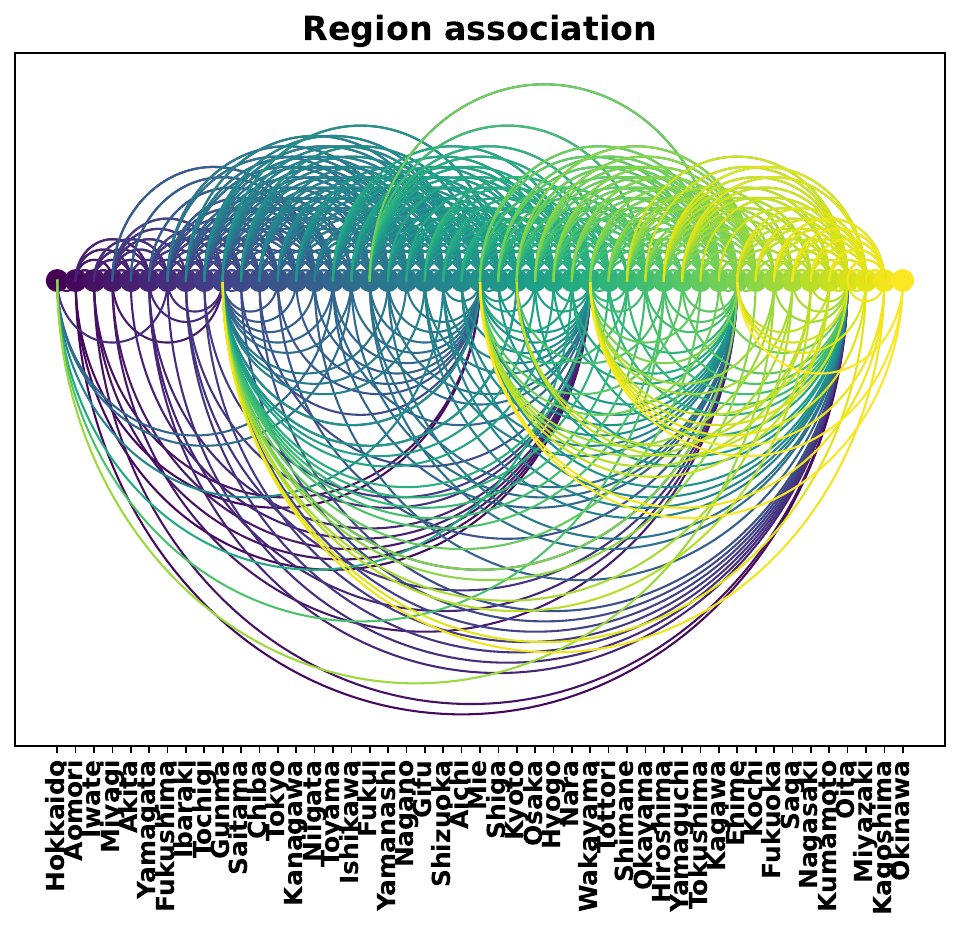}
    \caption{\label{fig:connection} Each dot represents a prefecture in Japan, and the prefectures are ordered from left to right based on how north they are. For connections above the dots, a link is derived if the distance between a pair of prefectures is less than 1000 km. For connections below the dots, a link is derived if the coverage probability of the forecasted curve of a prefecture is higher than 90\% for all other prefectures. The connections derived from coverage probabilities are directional whereas the top connections are not.}
\end{figure}

\section{Simulation Studies}\label{sec:simulation}

In simulation studies,
we will use the coverage probability to evaluate the forecast performance. The coverage probability is defined as the proportion of the true functional data within the $(1-\alpha)\%$ prediction interval. In addition, we will conduct a sensitivity analysis of the model to tuning parameters and investigate the scalability of the proposed \textit{ProFnet}.

\subsection{Simulation from Latent Gaussian Processes}\label{ssec:sim1}
The data simulation starts with sampling from a collection pre-specified Gaussian process. Let $K$ be the total number of latent Gaussian Processes (GPs) with a pair of coefficients $(\mu_{k}, \rho_{k})$, where each GP has a constant mean $\mu_{k}$ and exponentially squared kernel function with length-scale $\rho{k}$. The mean is generated based on $\text{N}(0, s^{2})$ and each length-scale is randomly generated by a $\text{Gamma}(a,b)$. We can generate random prior samples from the $K$ Gaussian processes for half of the $T$ points. Then, we can fill the remaining latent values from random samples of the posterior distribution of each Gaussian process. After obtaining $\mathbf{z}_{t}$ for all $t = 1,..., T$, we can use then simulate the actual high-dimensional functional time series. We use a random initialized embedding layer to encode the region information for all $h = 1,..., H$ and use a random initialized neural network to simulate individual functions at each $t$ and $h$ given  $\mathbf{z}_{t}$ and $W_{h}$. 

We examines how the model performs given different ratios of the number of time points $T$ and the number of regions $H$. We will also investigate the impact of the number ($K$) of latent Gaussian processes used in the model. The results are summarized in Table~\ref{tab:4}. We will fix $T = 50$ and only choose $H$ to adjust the ratio. We will split the data into training (first 80\% of time points) and testing (remaining 20\% of time points), and the coverage probability will be calculated based on the testing data set. Furthermore, we can see the forecast for a given pair from $t$ to $t^{\prime}$ at $h^{\prime}$ can originate for all $h = 1,..., H$. That is, \textit{ProFnet} is able to forecast the functional time series for all regions simultaneously which makes $H^{2}$ pairs of forecasts at a time. Table~\ref{tab:4} is presenting the coverage probability on test set after we pick the model has the best prediction performancefor training data.

\begin{table}[H]
    \caption{\label{tab:4} Simulation based on the latent Gaussian processes. The coverage probability of the proposed \textit{ProFnet} with different numbers of latent Gaussian processes used and different numbers of dimensions in functional time series. The coverage probability on testing data given best forecast model obtained from all pairs of $(h,h^{\prime})$ in model training step.}
    \begin{center}
        \begin{small}
        \begin{sc}
            \begin{tabular}{|c|c|c|c|c|c|}
                \hline
                \backslashbox{K}{H} & 10 & 20 & 50 & 100 & 500 \\ \hline
                8   & 0.95 & 0.99 & 0.96 & 0.97 & 1.00 \\ \hline
                16  & 0.89 & 0.98 & 1.00 & 1.00 & 1.00 \\ \hline
                32  & 0.99 & 1.00 & 1.00 & 1.00 & 1.00 \\ \hline
                64  & 1.00 & 1.00 & 1.00 & 1.00 & 1.00 \\ \hline
                128 & 0.99 & 1.00 & 1.00 & 1.00 & 1.00 \\ \hline
                256 & 1.00 & 1.00 & 1.00 & 1.00 & 1.00 \\ \hline
                512 & 1.00 & 1.00 & 1.00 & 1.00 & 1.00 \\ \hline
                1024& 0.99 & 1.00 & 1.00 & 1.00 & 1.00 \\ \hline
            \end{tabular}
\end{sc}
\end{small}
\end{center}
\vskip -0.1in
\end{table}

From Table~\ref{tab:4}, we can see that the coverage probability is increasing as the number of latent Gaussian processes $K$ increases. This is expected as the model is able to capture more complex patterns in the functional time series. We have to be cautious of overfitting when the number of latent Gaussian processes is too large. For example in Table~\ref{tab:4}, the coverage probability is decreasing when $K \geq 256$ and $H = 10$.

In addition to assessing the forecast ability of the model, we also investigate the scalability of our neural network. We will consider the computational time as a prime indicator and consider different scales of the modeling problem given the number of latent Gaussian processes. The results are summarized in the following Figure~\ref{fig:scalability}.
\begin{figure}[H]
    \centering
    \includegraphics[width=\textwidth]{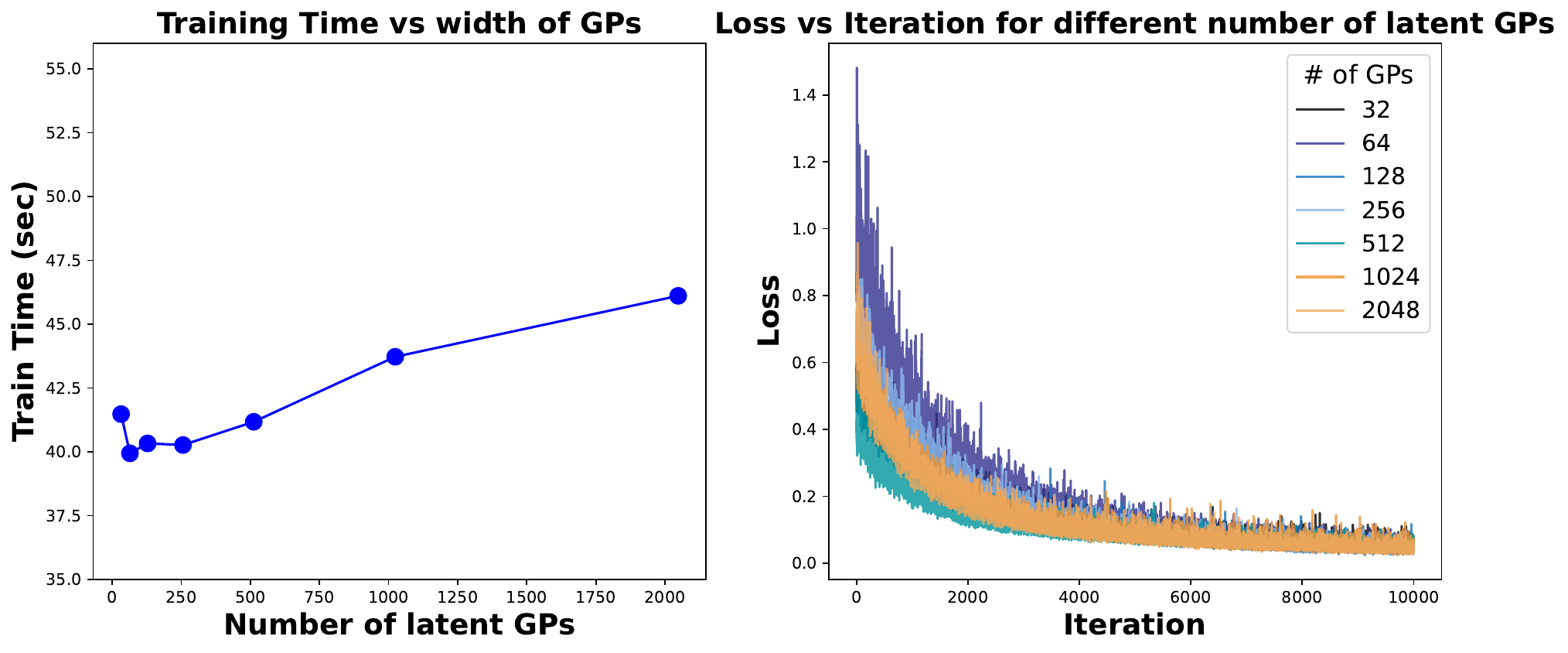}
    \caption{\label{fig:scalability} On the left, we show the computational time (per 10000 updates) of the proposed \textit{ProFnet} with different numbers of latent Gaussian processes used. The right plot shows the tracing plot of objective function value over the number of updates given different numbers of GPs used.}
\end{figure}
\vspace*{-0.5cm}
The training time is calculated as time required for running 10000 stochastic gradient updates. We can see the training time is linearly increasing as the number of latent GPs increases. Furthermore, the tracing plot of the objective function value over the number of updates is shown in the right plot of Figure~\ref{fig:scalability}. The model is converging and the convergence is faster or indifferent to a larger number of latent GPs.

\section{Conclusion}\label{sec:conclusion}

In this work, we propose a probabilistic neural network for high-dimensional functional time series, \textit{ProFnet}. The proposed model is able to provide a probabilistic forecast for high-dimensional functional time series and quantify the forecast uncertainty. The model is built based on the feedforward neural network structure which is more favorable than the RNN structure. The model is also flexible in that it can be easily extended to deep learning and can be used for high-dimensional and large-scale functional time series data. The proposed model is also able to handle the trend and nonstationarity of functional data which is not possible in the traditional FDA methods. The model is designed to forecast different lags simultaneously. The proposed model is evaluated through simulation studies and an application in forecasting Japan's mortality rate curves. The results show that the proposed \textit{ProFnet} is superior in forecasting future data and is computationally feasible and scalable to high-dimensional functional time series data.

\bibliographystyle{apalike}
\bibliography{profnet}

\end{document}